\documentclass{article}

\usepackage{widetable}
\usepackage{siunitx}
\usepackage{amsfonts} 
\usepackage{amsmath} 
\usepackage{xcolor}
\usepackage{booktabs}
\usepackage{caption} 
\usepackage{placeins}

\usepackage[utf8]{inputenc} %
\usepackage[T1]{fontenc}    %
\usepackage{hyperref}       %
\usepackage{url}            %
\usepackage{booktabs}       %
\usepackage{amsfonts}       %
\usepackage{nicefrac}       %
\usepackage{microtype}      %
\usepackage{xcolor}         %
\usepackage{amsmath,amsthm,amssymb}

\usepackage{subcaption}

\usepackage[capitalise]{cleveref}  %
\usepackage{comment}
\usepackage[inline]{enumitem}

\usepackage{graphicx}
\usepackage{grffile}
\usepackage{wrapfig,epsfig}
\usepackage{epstopdf}
\usepackage{algpseudocode}
\usepackage{multirow}
\usepackage[T1]{fontenc}
\usepackage{bbm}
\usepackage{comment}
\usepackage{dsfont}
\usepackage{makecell}
\usepackage{enumitem}
\usepackage{booktabs}

\usepackage{amsmath}
\usepackage{amsthm}
\usepackage{amssymb}
\usepackage{algorithm}
\usepackage{color}
\usepackage[english]{babel}
\usepackage{graphicx}
\usepackage{grffile}
\usepackage[numbers,sort]{natbib}
\usepackage{wrapfig,epsfig}
\usepackage{epstopdf}
\usepackage{algpseudocode}
\usepackage{multirow}
\usepackage[T1]{fontenc}
\usepackage{bbm}
\usepackage{comment}
\usepackage{dsfont}
\usepackage{makecell}
\usepackage{enumitem}
\usepackage{booktabs}
\usepackage{afterpage}

\usepackage{minitoc}

\usepackage{algorithm}
\usepackage{color}
\usepackage[english]{babel}

\usepackage{import}

\newlength{\defbaselineskip}
\newcommand{\norm}[1]{\left\lVert#1\right\rVert_2}

\newcommand\extrafootertext[1]{%
    \bgroup
    \renewcommand\thefootnote{\fnsymbol{footnote}}%
    \renewcommand\thempfootnote{\fnsymbol{mpfootnote}}%
    \footnotetext[0]{#1}%
    \egroup
}
\newcommand{\COtwo}{CO$_2$ }
\newcommand{\COtwoNOSPACE}{CO$_2$}

\title{Latent Space Simulation for Carbon Capture Design Optimization}
    \usepackage{authblk}
\author{
    Brian Bartoldson,\textsuperscript{\rm 1}
    Rui Wang,\textsuperscript{\rm {1,2}}
    Yucheng Fu,\textsuperscript{\rm 3}
    David Widemann,\textsuperscript{\rm {1}} \\
    Sam Nguyen,\textsuperscript{\rm 1}
    Jie Bao,\textsuperscript{\rm 3} 
    Zhijie Xu,\textsuperscript{\rm 3}
    Brenda Ng\textsuperscript{\rm 1}
}
\affil{
    \textsuperscript{\rm 1}Lawrence Livermore National Laboratory; \textsuperscript{\rm 2}UCSD;
    \textsuperscript{\rm 3}Pacific Northwest National Laboratory \\
    {\normalsize \texttt{\{bartoldson, widemann1, nguyen116, ng30\}@llnl.gov;
    ruw020@ucsd.edu; 
    \{yucheng.fu, jie.bao, zhijie.xu\}@pnnl.gov}}
}

\begin{document}

\maketitle

\begin{abstract}
The \COtwo capture efficiency in solvent-based carbon capture systems (CCSs) critically depends on the gas-solvent interfacial area (IA), making maximization of IA a foundational challenge in CCS design. While the IA associated with a particular CCS design can be estimated via a computational fluid dynamics (CFD) simulation, using CFD to derive the IAs associated with numerous CCS designs is prohibitively costly. Fortunately, previous works such as Deep Fluids (DF) (Kim et al., 2019) show that large simulation speedups are achievable by replacing CFD simulators with neural network (NN) surrogates that faithfully mimic the CFD simulation process. This raises the possibility of a fast, accurate replacement for a CFD simulator and therefore efficient approximation of the IAs required by CCS design optimization. Thus, here, we build on the DF approach to develop surrogates that can successfully be applied to our complex carbon-capture CFD simulations. Our optimized DF-style surrogates produce large speedups (4000x) while obtaining IA relative errors as low as 4\% on unseen CCS configurations that lie within the range of training configurations. This hints at the promise of NN surrogates for our CCS design optimization problem. Nonetheless, DF has inherent limitations with respect to CCS design (e.g., limited transferability of trained models to new CCS packings). We conclude with ideas to address these challenges. 
\end{abstract}

\section{Introduction}
\extrafootertext{This is an extended version of a paper appearing in the Proceedings of the 34th Annual Conference on Innovative Applications of Artificial Intelligence (IAAI-22).}
Reducing the carbon intensity of electricity generation, the leading contributor of global greenhouse gas (GHG) emissions, is a key component of work towards achieving stabilization of GHG concentrations at a safe level \citep{edenhofer2015climate}. Critically, GHG emissions of fossil-based power plants can be reduced via carbon dioxide (\COtwoNOSPACE) capture technologies \citep{edenhofer2015climate}. 

\begin{figure}[t]
\centering
\begin{subfigure}[b]{.33\columnwidth}
    \centering
    \includegraphics[width=1\columnwidth]{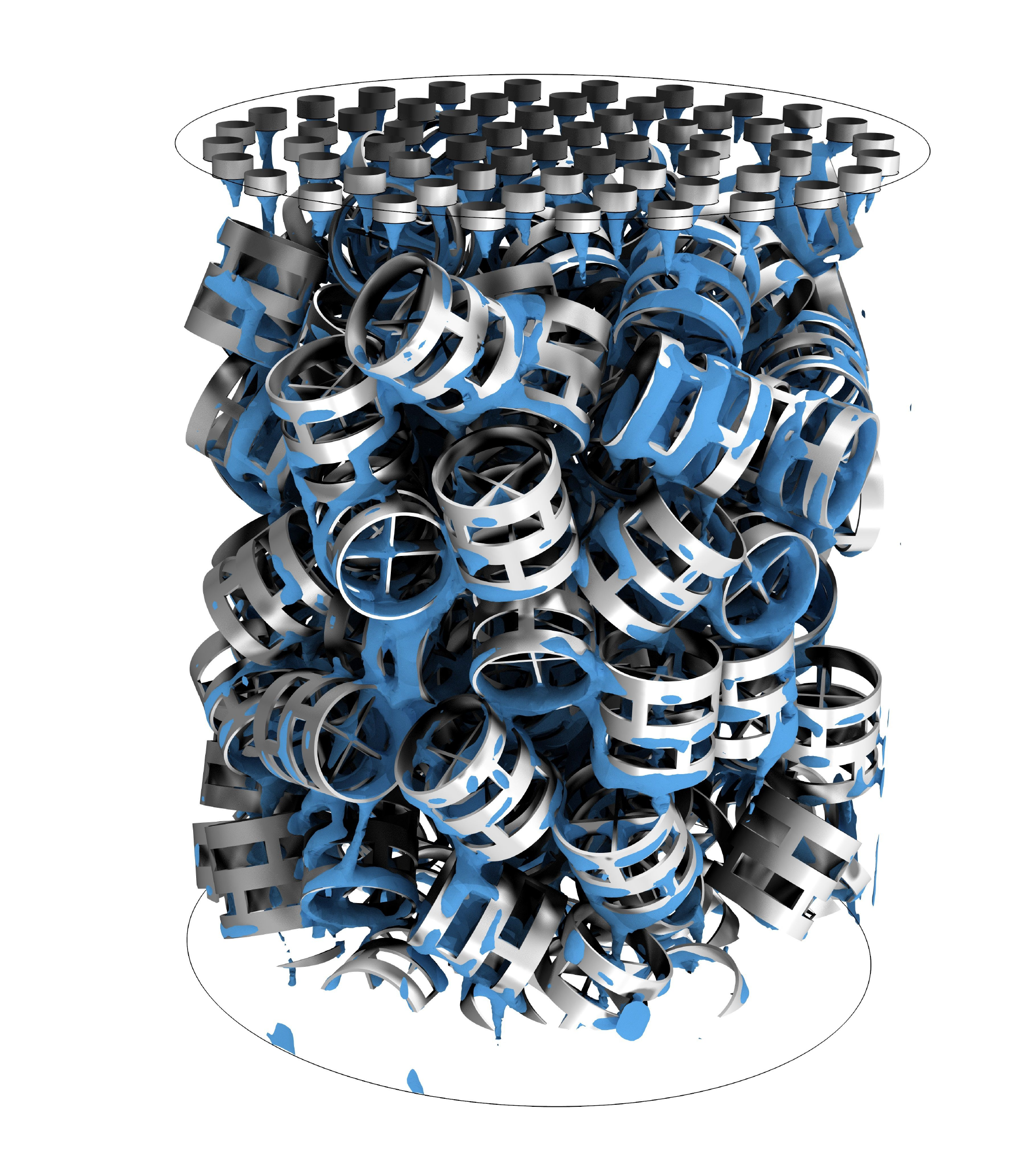} 
    \caption{3D CCS RCM}
    \label{fig:RCM}
\end{subfigure}%
\hspace*{15mm}
\begin{subfigure}[b]{.33\columnwidth}
    \centering
        \includegraphics[width=1\columnwidth]{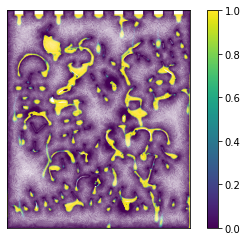}
    \vspace*{1mm}
    \caption{2D volume fraction field}
    \label{fig:2D}
\end{subfigure}
\begin{subfigure}{.5\columnwidth}
    \centering
    \vspace*{2mm}
    \includegraphics[width=1\columnwidth]{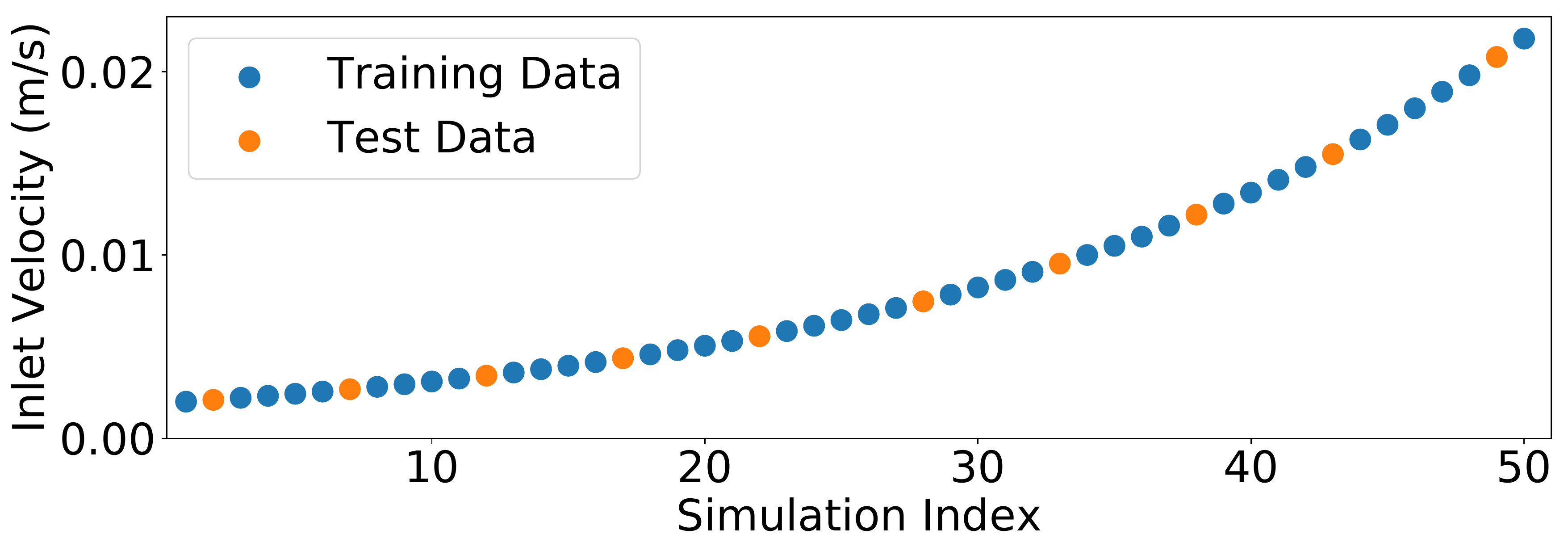} 
    \caption{Solvent inlet velocity $v$ for each CFD simulation.}
    \label{fig:interpolate}
\end{subfigure}
\caption{(a) The 3D RCM of our CCS. In our simulation, parameter $v$ is the rate at which CO$_2$BOL (solvent) is injected into the column from above, while \COtwoNOSPACE-laden gas is injected from below. (b) Volume fraction field in 2D domain at $t=500$ and $v=\SI{0.01}{\meter/\second}$. (c) The 50 CCS configurations we used to train and test our DF-inspired surrogate.}
\label{fig:data}
\end{figure}

Capture of \COtwo from power-plant flue gas is generally achieved through pre-combustion, oxyfuel-combustion, or post-combustion technologies \citep{koytsoumpa2018co2}. Of these, the most widely adopted is the solvent-based post-combustion approach \citep{koytsoumpa2018co2,wang2017review}, wherein \COtwo is captured through an absorption process caused by interaction between a particular liquid solvent and the flue gas inside a reactor column filled with packings (see Figure \ref{fig:RCM}). 

A foundational challenge underlying design of such solvent-based carbon capture systems (CCSs) is optimizing the pairing of solvent and packing to maximize the gas-solvent interfacial area (IA) $a$ for the \COtwo absorption reaction. Notably, \COtwo capture efficiency is determined by $k_ga$, where $k_g$ is the gas film mass transfer coefficient \citep{singh2017hydrodynamics, song2018mass}.

Capturing local effects from the hydrodynamics, heat, and mass transfer of the absorption process, CFD simulation helps us understand the IA associated with a solvent and packing configuration. While physically faithful, CFD is too computationally expensive to be used to evaluate potential configurations inside a CCS design optimization process.

We address this challenge by learning neural network (NN) surrogates for our CFD simulator. Specifically, we apply the Deep Fluids \citep{kim2019deep} surrogate approach and variants we introduce to CFD-simulated CCS volume fraction fields (from which we can compute IA). Once trained, these surrogates can quickly simulate the volume fraction fields of an unseen CCS configuration, enabling fast IA predictions for our CCS design optimization process.

Critically, our experiments focus on whether these fast surrogates accurately predict IA and liquid volume fraction. 

While the original Deep Fluids (DF) approach gave visually realistic results on simpler problems, it produced high errors on our CCS application's complex domain. Thus, we developed multiple innovations to customize DF (e.g., end-to-end training and new NNs for latent-space simulation). Our innovations led to a final DF-inspired surrogate that can simulate CCSs with unseen configurations and predict IA with 4\% relative error compared to the CFD simulation that is 4000x slower. This result demonstrates the promise of NN-based simulation for CCS design optimization.

\section{Relevance of AI to Application}
Modeling the spatiotemporal dynamics of a system is essential not just to our CCS application, but to a wide array of applications in physics, epidemiology, and molecular dynamics. Traditional approaches rely on running numerical simulations: known physical laws encoded in systems of complicated coupled differential equations are solved over space and time via numerical differentiation and integration schemes. However, such methods are computationally intensive, requiring expertise and manual engineering in each application \cite{Houska2012, 
Butcher1996numerical}. 

Consequently, there is an active, expansive literature on deep learning (DL) methods for accelerating or replacing numerical simulations \cite{Wang2021PhysicsGuidedDL, 
Willard2020Survey}. For example, deep dynamics models can approximate high-dimensional spatiotemporal dynamics by directly forecasting future states, bypassing numerical integration \cite{
Wang2020TF, 
Wang2020symmetry, Bezenac2018Deep}. 
\citet{SanchezGonzalez2020LearningTS} designed a deep encoder-processor-decoder graph architecture for simulating fluid dynamics under a Lagrangian description.
\citet{BelbutePeres2020CombiningDP} combined graph neural networks with a CFD simulator run on a coarse mesh to generate high-resolution fluid flow predictions.
\citet{tompson2017accelerating} replaced the numerical pressure solver with convolutional networks in Eulerian fluid simulation and obtained realistic results.

When a CFD model is replaced with a surrogate that forecasts future states within a lower-dimensional latent space, larger speedups are realizable. These 
``latent space physics'' can be learned with an autoencoder that maps physical fields to latent vectors and an LSTM that models the temporal evolution of the latent vectors. These surrogates enable fluid flow pressure field simulations hundreds of times faster than CFD's \citep{steffen2019latent}. In Deep Fluids (DF) \cite{kim2019deep}, the LSTM is replaced by an MLP, and an auxiliary term is added to the loss to ensure divergence-free motion for incompressible flows.

The speedup provided by these methods is relevant to our CCS design optimization, which requires volume fraction field simulations for numerous CCS configurations. Accordingly, we applied DF to our CCS data (see Section \ref{sec:approach}). Our innovations for DF include new latent-space simulation networks---we explore various MLPs, CNNs, LSTMs \citep{hochreiter1997long}, and transformers \citep{Vaswani2017AttentionIA}---as well as end-to-end training.

\section{Application and AI Approach}
\label{sec:approach}

This work augments DF to achieve acceptable surrogate predictions for our \COtwoNOSPACE-capture-simulation application. Section \ref{sec:data} describes our application and its related CFD data. Section \ref{sec:surrogates} discusses our DF-inspired AI surrogates and the innovations we developed to improve DF's performance.

\subsection{Application and CFD Data Description}
\label{sec:data}
Our application is IA-based design optimization of CCSs. In particular, we seek a means to find CCSs associated with high IA, a key determinant of \COtwoNOSPACE-capture efficiency. A typical (but slow) approach is to predict IA using the volume fraction fields resulting from a CFD simulation of a \emph{representative column model} (RCM) (Figure \ref{fig:RCM}), which is a representative section of the bench-scale column of our CCS.

This involves simulating the counter-current solvent and gas flow in the 3D RCM (containing 6.5 million data points) then solving continuity and momentum equations with the Volume of Fluid (VOF) method \citep{fu2020investigation}:

\begin{equation}
\frac{\partial \rho}{\partial t}+\nabla \cdot \rho \mathbf{u}=0,
\end{equation}
\begin{equation}
\frac{\partial(\rho \mathbf{u})}{\partial t}+\nabla \cdot(\rho \mathbf{u u})=-\nabla p+\mu \nabla^{2} \mathbf{u}+\rho \mathbf{g}+\mathbf{F}_{\sigma},
\end{equation}
where $\rho$ is density, $\mu$  is viscosity, $\mathbf{u}$ is velocity, $p$ is pressure, $\mathbf{g}$ is gravity, and $\mathbf{F}_{\sigma}$ is the surface tension arising at the gas-liquid interface. The density and viscosity are calculated by a volume fraction average of liquid ($\alpha$) and gas phase (1-$\alpha$). The evolution of $\alpha$ is governed by the transport equation:
\begin{equation}
\frac{\partial \alpha}{\partial t}+\nabla \cdot (\mathbf{u} \alpha)=0.
\end{equation}

Running many such CFD simulations to thoroughly explore the CCS configuration space (which involves: packing type, e.g., Gyroid vs. Schwarz-D; inlet velocity $v$; solvent viscosity/surface-tension; etc.) is computationally infeasible. However, it may be possible to use existing CFD data to train an AI surrogate model to quickly and accurately simulate dynamics of CCSs with never-before-seen configuration settings. During IA-based optimization, this surrogate's faster speed would allow more thorough exploration of the CCS configuration space, improving the chance of finding a CCS design with better \COtwoNOSPACE-capture performance.

To evaluate the utility of AI surrogates for our application, we measure the speed and accuracy of surrogate simulations of CCSs with values of the solvent inlet velocity parameter not seen during training. Our data is $N=50$ CCS CFD simulations, each with a unique inlet velocity (Figure \ref{fig:interpolate}). $N_{train}=40$ simulations are used for training, while the remaining $N_{test}=10$ simulations are for evaluation.

As a first step, we worked with CFD data from a 2D domain, derived from a vertical slice of the larger 3D CCS domain (Section \ref{sec:deploy} discusses extension to 3D). Like the 3D domain, the 2D domain with 150,073 polygonal mesh grid points  has its counter-current flow solved via the VOF method. Starting from an initial state, a particular CCS configuration on this 2D domain reaches a steady state in 5000 timesteps (Figure \ref{fig:2D} shows a single timestep), using a wall-clock time $W_{\mathrm{CFD}} = \SI{3600}{\second}$ when run on 96 CPU cores. 

Before training surrogates, the simulation data is temporally downsampled to 500 timesteps. As our AI approach uses CNNs that operate on uniform grids, the simulation data is then spatially downsampled via linear interpolation to obtain uniform grid data $g_t^{(i)} \in G$, where $G = \mathbb{R}^{k \times k}$, $i \in {1,2,...,50}$ denotes the simulation index, and $t \in \{1,2,...,T\}$ is the timestep with $T=500$ (we omit $i$ and $t$ when referring to generic elements of $G$). We report results for $k=64$ but saw similar performance with $k=128$.

Note that we train surrogate models only on volume-fraction field dynamics because our application is focused on IA, which is a function of the volume-fraction field $g_t^{(i)}$.

\subsection{AI Approach to Our CFD Simulations}
\label{sec:surrogates}

\begin{figure*}[t]
\centering
\includegraphics[width=1\textwidth]{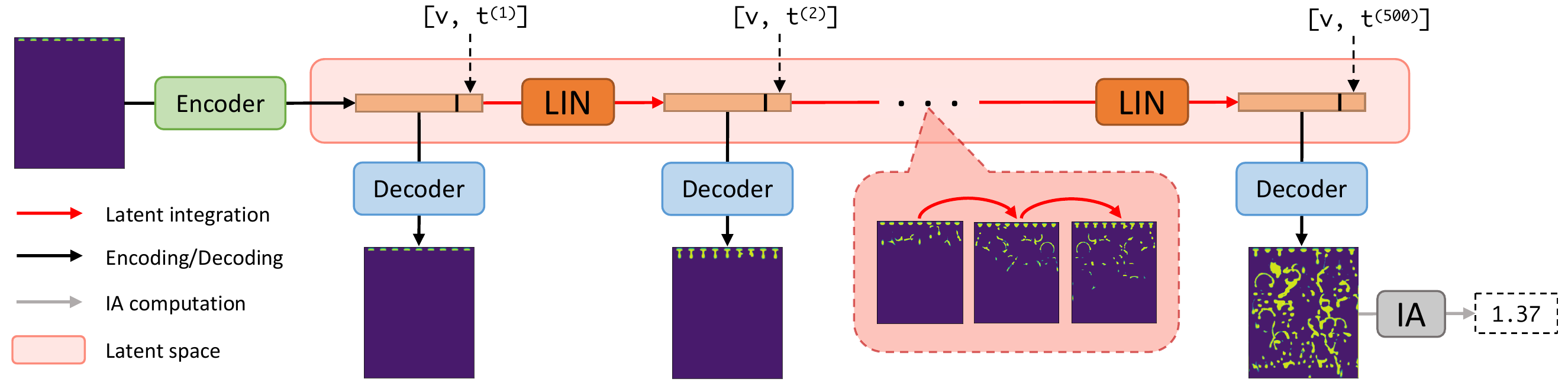} 
\caption{DF surrogate simulation (``full rollout'', $w=T-1$) of CCS's 2D volume fraction field, with IA computation at $t=T$.}
\label{fig:NNs}
\end{figure*}

Here, we discuss surrogate simulation with DF and our innovations. DF is a data-driven, NN-based, non-intrusive surrogate approach. In our application, surrogate input is the timestep $t \in \mathbb{N}$, a \emph{frame} $g_t \in G$ that represents the volume fraction field at that timestep, and the inlet velocity $v \in \mathbb{R}$ of the CCS. Surrogate outputs $\widehat{g}_{t+j}$ are volume fraction fields at times $t+j$. (Note, we use hats to denote predicted quantities.) DF achieves this as follows (see Figure \ref{fig:NNs}): 

\begin{enumerate}
    \item \textbf{Encode the frame:} Reduce the dimensionality of the frame $g_t$ via an \emph{encoder network} $E: G \rightarrow \mathbb{R}^c$, where $c$ is the latent space's dimension. The output of $E$ is a \emph{latent vector} denoted by $l_t$.
    \item \textbf{Add simulation-configuration details to latent vector:} Replace the last two elements of $l_t=E(g_t)$ with $[v,t]$. 
    \item \textbf{Temporally advance the latent vector:} Use a separate \emph{latent integration network} (LIN) $I: \mathbb{R}^c \rightarrow \mathbb{R}^c$ to predict the change $\Delta l_t = l_{t+1}-l_t$ and thereby obtain the next latent vector $\widehat{l}_{t+1}=l_t + I(l_t)$; perform this step $w$ times.
    \item \textbf{Decode the new latent vector(s):} Feed $\widehat{l}_{t+j}$ to the \emph{decoder network} $D:\mathbb{R}^c \rightarrow G$ to obtain $\widehat{g}_{t+j}$.
\end{enumerate}

$E$ and $D$ are trained together as an autoencoder, more generally as a latent vector model $\mathrm{LVM}: G \rightarrow G \times \mathbb{R}^c$, which encodes then decodes a $g_t$ and provides the corresponding latent vector (or encoding) $l_t$ as output. Notably, by inserting the simulation-specific information $[v,t]$ into the encoder's latent code rather than predicting it, we remove the supervised learning of these values done in \citep{kim2019deep} in exchange for allowing the decoder and LIN to train with the exact simulation parameters $[v,t]$. Given a trained LVM, the DF approach computes the latent vector $l$ for every training simulation frame $g$. The LIN $I$ is then trained to predict the deltas among these $l$ within a \emph{rollout} window  ($\Delta l_\tau, \tau \in \{t,...,t+w-1\}$ with $w$ as the window size) given only $l_{t}$ (and thus also $[v,t]$) as input. 

Once $I$ is trained, it can be iteratively applied to perform a full rollout that produces $\widehat{l}_{t}$ and  $\widehat{g}_t$ for $t \in \{1,...,T\}$, given only $g_1$ and $v$ as input. Full rollouts are useful when using trained surrogates to replace CFD simulation (see Figure \ref{fig:NNs}).

In Section \ref{sec:experiments}, we study how surrogate performance changes in response to modifying the training-time rollout window size $w$ and other variables. Additionally, we evaluate larger modifications to the DF approach: joint training of the LVM and LIN (end-to-end training), and new LVM/LIN architectures (e.g., transformers).

\paragraph{Latent vector models} Like the original DF approach, we use a CNN autoencoder (comprising both $E$ and $D$) as an LVM. The encoder network $E$ is four Conv2D layers  (with stride 2, padding 1, kernel size 3, and layer-specific channel counts $[128, 256, 512, 1024]$) followed by a linear layer that maps from the final convolutional feature map to $\mathbb{R}^c$. The decoder network $D$ completes a symmetric architecture for the LVM, with a linear layer followed by four ConvTranspose2D layers (using the layer-specific channel counts $[512, 256, 128, 1]$) so that the output has $g$'s shape.

We also consider replacing the CNN autoencoder with an LVM derived from an SVD of the training data ($B=U\Sigma V^\top$, where $B\in \mathbb{R}^{k^2\times N_{train}T}$). Specifically, encoding then decoding $B$ via multiplication with the truncated matrix $U_{\cdot,1:c}$ gives the best rank-$c$ approximation $\hat{B}$ \citep{eckart1936approximation}. This provides a strong baseline for the larger and slower CNN LVM, which also compresses frames to size $c$ but uses nonlinear transformations to make possible decodings $\hat{B}$ with rank $m{>}c$ and thus smaller error.

Finally, we explore the usage of vision transformers (ViTs) as LVMs. As in \citet{dosovitskiy2021an}, we split each input frame into 16$\times$16 patches, linearly embed each of them, add position embeddings, and feed the resulting sequence of vectors to a standard transformer encoder. Then, the transformer decoder takes as input the latent vectors from the encoder and an extra learnable embedding. The learnable embedding's state at the output of the transformer decoder serves as the frame representation and is linearly transformed to generate the reconstructed frame.


\paragraph{Latent integration networks}
As in DF, we use an MLP to learn $I$, the latent integration network (LIN). We explore several modifications to the proposed MLP LIN. We modify the LIN's depth and width, and augment the input dimension to exploit more history (i.e., more frames as input) such that $I : \mathbb{R}^{s\times c} \rightarrow \mathbb{R}^c$, where sequence length $s$ is the number of prior timesteps the LIN uses as input (when $t<s$, the $s-t$ remaining rows/timesteps of the input space are initialized at $\mathbf{0}$ and filled gradually as the LIN produces them).

In addition to MLPs, we evaluated other competitive sequence models, such as CNNs \citep{oord2016wavenet}, LSTMs \citep{hochreiter1997long}, and transformers  \citep{Vaswani2017AttentionIA}, as LINs. For the LSTM and transformer networks, we use PyTorch implementations: the transformer is a series of six 8-head transformer encoder layers with hidden dimension $c$, while the LSTM network is a series of LSTM layers with hidden dimensions $H=[h_1,...,h_n]$. To test CNNs, we created AutoRegressive CNN (ARC) LINs that comprise 1D convolutional layers with output channel counts $[h_1, h_2, ..., h_n, c]$, the first layer using $c$ input channels, and all layers using kernel size 3 and padding 1. The output layer of each LIN is a linear layer that maps to $\mathbb{R}^c$, except for the ARC LIN's output layer, which uses a learnable linear combination of the rows from the final feature map to compute $\Delta l_t \in \mathbb{R}^c$. Finally, with the ViT LVM, we used a single linear layer for the LIN, as the performance bottleneck when using the ViT was the LVM.

\paragraph{Performance optimization and evaluation}
When training LVMs, we optimize the relative error loss $L_{\mathrm{RE}}(\widehat{g},g) = \frac{\norm{\widehat{g}-g}}{\norm{g}}$ averaged across the $K$ volume fraction fields in a batch, as we found this performed better than RMSE and preliminary results showed no quantitative benefit of the gradient-matching loss term used in DF. Similarly, to train the LIN, we optimize the relative error loss $L_{\mathrm{RE}}(\widehat{l}_t,l_t)$ averaged over the $K$ $w$-timestep latent vector rollouts (i.e., there are $Kw$ unique relative errors each batch). When using end-to-end training, the $Kw$ rolled-out latent vectors are decoded to frames $\widehat{g}_t$, then $L_{\mathrm{RE}}(\widehat{g_t},g_t)$ is computed and gradients are taken with respect to the parameters of $E$, $I$, and $D$.
We show the effect of training the LIN with RMSE (instead of $L_{\mathrm{RE}}$) and with different values of $w$ and $s$ in Section \ref{sec:sweep}.

While our loss functions aim to ensure accurate reconstruction of $g_t$, our ultimate goal is IA-based optimization of CCS designs. Thus, we are primarily interested in the ability of DF-inspired surrogates to estimate steady-state IA correctly, given only the initial conditions $g_1$ and inlet velocity $v$ of a new, unseen CCS. After training our models, we measure the steady-state IA relative error $
L_{\mathrm{RE}}\left(IA(\widehat{g}_{T}^{(i)}), IA(g_{T}^{(i)})\right),$
where $i$ indexes the $N_{test}$ previously-unseen test simulations, $\widehat{g}_{T}^{(i)}$ is derived from a full rollout, and $IA : G \rightarrow \mathbb{R}$ is the integral of surface areas for which the volume fraction is 50\% (on a 2D volume fraction field, such surfaces correspond to volume fraction isocurves).

Then, we report the mean error: $\mathrm{Error}_{\mathrm{IA}} = \frac{1}{N_{test}} \sum_{i=1}^{N_{test}} L_{\mathrm{RE}}\left(IA(\widehat{g}_{T}^{(i)}), IA(g_{T}^{(i)})\right)$.

While this mean, final-timestep IA relative error is a critical metric for our application, we are also interested in whether surrogates could potentially replace CFD simulations for purposes other than IA estimation. Thus, to test the ability of surrogates to mimic CFD volume fraction dynamics at all timesteps, we use $\widehat{g}_{t}^{(i)}$ from full rollouts on the test data to compute average volume fraction relative error: $\mathrm{Error}_{\mathrm{VF}} = \frac{1}{N_{test}T} \sum_{i=1}^{N_{test}} \sum_{t=1}^{T} L_{\mathrm{RE}}\left(\widehat{g}_{t}^{(i)}, g_{t}^{(i)}\right)$. 

Finally, we use the ratio of the wall-clock time of the CFD simulation to that of a surrogate's full rollout (which is run on four NVIDIA P100 GPUs) to compute the relative speedup offered by our AI approach: $S_W = \frac{W_{\mathrm{CFD}}}{W_{\mathrm{AI}}}$. 

\section{Experiments}
\label{sec:experiments}
To understand whether AI surrogates may help our IA-based CCS-design-optimization task, we use our CFD data to study the accuracy and speedups of DF surrogates and our innovative approaches. We report the mean of two runs (with standard deviation in parentheses) for each result. The code we use to run our experiments is available here: \url{https://github.com/CCSI-Toolset/DeeperFluids}.

The performance of DF-inspired surrogates depends on the performances of both LIN ($I$) and LVM ($E$ and $D$). So we first study performance of LVMs (Section \ref{sec:lvm_comparison}). Given the best LVM, we use its latent vectors to train LINs, with which we perform ``full rollouts'' to evaluate performance of the entire surrogate system via $\mathrm{Error}_{\mathrm{VF}}$, $\mathrm{Error}_{\mathrm{IA}}$, and $S_W$ (Section \ref{sec:lin_comparison}). Section \ref{sec:e2e} then tests the benefit of end-to-end training of the LVM and LIN. Finally, we discuss the effects of key model/training hyperparameters (Section \ref{sec:sweep})

\subsection{LVM Architecture Comparison}
\label{sec:lvm_comparison}
To test how the DF approach might benefit our application, we first consider LVM error of three model classes, including two new approaches that we introduced to DF. Specifically, we compare the performances of a CNN-based autoencoder (used in DF); a ViT \citep{dosovitskiy2021an} with similar parameter count; and SVD, a faster, linear approach.

\begin{figure}[t]
    \centering
        \includegraphics[width=1\columnwidth]{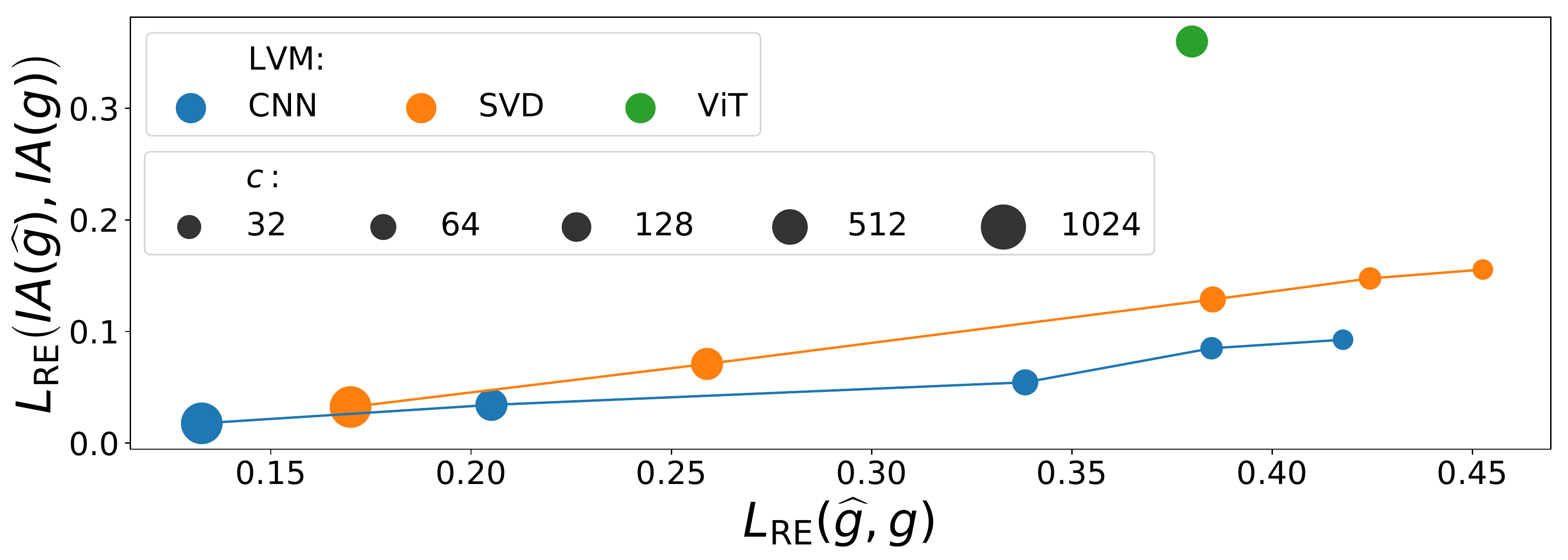}
\caption{Mean test error for our LVMs at up to five $c$ values.}
\label{fig:LVM}
\end{figure}

\sloppy To measure performance, we average the frame reconstruction and IA errors, $L_{\mathrm{RE}}(\widehat{g},g)$ and $L_{\mathrm{RE}}(IA(\widehat{g}_{T}), IA(g_{T}))$, across the test data. As SVD is linear, we expect it to be less accurate than CNNs/ViTs, which are powerful approximators that can represent nonlinear functions. However, ViTs and CNNs require careful optimization to perform well. This is especially true for ViTs, which lack the inductive biases of CNNs (e.g., translation equivariance and locality), harming their performance when data is scarce  \citep{dosovitskiy2021an} as it is here. Thus, SVD may potentially outperform CNNs or ViTs. Consistent with the difficulty of training ViTs without extra/sufficient data, we found SVD outperformed our ViT on the LVM reconstruction task  (Figure \ref{fig:LVM}). However, SVD did not outperform the CNN. The ability of CNNs but not ViTs to outperform SVD reinforces the (particularly important in our low-data regime) point that image-processing inductive biases provided by convolutional layers offer a significant benefit. As \textbf{CNN-based autoencoders produced the best LVMs}, we use them in our remaining experiments except where we indicate use of a ViT LVM.

\subsection{LIN Architecture Comparison}
\label{sec:lin_comparison}

Given the latent vectors from the CNN LVM, we sought to determine the best LIN architecture. While the original DF used an MLP for $I$, other architectures such as LSTM, CNN, and transformer offer competitive performance in sequence modeling. To test whether these other architectures may outperform MLPs on our data, we trained each architecture using several settings for $c$, $s$, $w$, $b$ (batch size), $H$, and the loss function (see Section \ref{sec:sweep} to see the effect of some of these factors). Then, with this set of trained models, we compared the best $\mathrm{Error}_{\mathrm{VF}}$ and $\mathrm{Error}_{\mathrm{IA}}$ of each architecture. 

\begin{table}[ht]
\centering
\begin{widetabular}{\columnwidth}{lccc}
\toprule
   LIN &  $\mathrm{Error}_{\mathrm{IA}}$  &  $\mathrm{Error}_{\mathrm{VF}}$  & $S_W$\\
\midrule
ARC &0.07 (0.00) &0.49 (0.00) &4800 \\ LSTM &0.08 (0.01) &0.49 (0.01) &2700 \\ MLP & \textbf{0.06} (0.00) & \textbf{0.47} (0.00) &5400 \\ Transformer &0.08 (0.04) &0.51 (0.00) &4300 \\
\bottomrule
\end{widetabular}
\caption{For each LIN architecture, the best $\mathrm{Error}_{\mathrm{IA}}$ and $\mathrm{Error}_{\mathrm{VF}}$ found via hyperparameter sweeps (standard deviations in parentheses) with wall-clock time speedup $S_W$.}
\label{tab:LIN}
\end{table}

While an MLP had the best performance across multiple runs ($0.06$ mean $\mathrm{Error}_{\mathrm{IA}}$), we found that  \textbf{the best single model for our IA-based optimization application was a transformer}, with $\mathrm{Error}_{\mathrm{IA}}$ of just $0.04$ (Table \ref{tab:LIN}). Each architecture in Table \ref{tab:LIN} has roughly \SI{1e5} parameters. For each performance measurement, the best models were found by optimizing $L_{\mathrm{RE}}$ rather than RMSE, except for the best transformer according to $\mathrm{Error}_{\mathrm{IA}}$, which was found by optimizing RMSE. In addition, the best models trained with $s=6$, except for the the best transformer and MLP models according to $\mathrm{Error}_{\mathrm{IA}}$, which trained with $s=1$. Finally, all models attained their best performance with $50\leq w \leq 300$.

The surrogate speedups are significant (up to 5400x) and partly stem from their 10x larger timesteps (see Section \ref{sec:data}). 

\subsection{End-to-End Training/Fine-tuning}
\label{sec:e2e}

The original DF trained LIN and LVM separately. Here, we investigate whether end-to-end (E2E) training improves performance. 
We applied E2E fine-tuning to a pretrained CNN LVM + ARC LIN surrogate and E2E training to a ViT LVM + linear LIN surrogate, then compared against classic training (Table \ref{tab:e2e}). Consistent with the benefit of LIN training that directly optimizes $L_{\mathrm{RE}}(D(\widehat{l}),g)$, we found both \textbf{E2E approaches significantly improved $\mathrm{Error}_{\mathrm{VF}}$ and $\mathrm{Error}_{\mathrm{IA}}$}.

\begin{table}[ht]
\centering
\begin{widetabular}{\columnwidth}{lcccc}
\toprule
 Surrogate & Training style &  $\mathrm{Error}_{\mathrm{IA}}$  &  $\mathrm{Error}_{\mathrm{VF}}$  \\
\midrule
CNN+ARC & Classic DF &                0.07 (0.00) &        0.52 (0.01)	 \\
CNN+ARC & E2E DF &                \textbf{0.06} (0.00) &        \textbf{0.48} (0.00) \\
\midrule
ViT+Linear & Classic DF &            0.53 (0.25)   &    0.77 (0.08) \\
ViT+Linear & E2E DF &                0.36 (0.08) &     0.59 (0.03)   \\
\bottomrule
\end{widetabular}
\caption{The effect of end-to-end training/fine-tuning.}
\label{tab:e2e}
\end{table}

\subsection{Hyperparameter Study}
\label{sec:sweep}

Here, we illustrate the effects of key DF hyperparameters, noting when optimal settings deviate from those in the original DF model. Settings of factors not shown in a table are held constant and are discussed in Appendix \ref{sec:appendix}. 

\paragraph{Deeper LINs and larger rollout windows have higher performance} We found that, for our data, the wide MLP LIN with $H=[1024,512]$ used in the original DF performed worse than a narrower, deeper MLP LIN with $H=[128,128, 128]$ (Table \ref{tab:w}). Since the narrower, deeper LIN offered superior performance with far fewer parameters, we used  $H=[128,128, 128]$ for our other experiments. Additionally, we found that increasing the window size, which allows the model to see and learn to control accumulation of error from repeated LIN applications during training, is helpful up to a point: as shown by the high error at $w=499$, too many LIN applications make training difficult.

\begin{table}[ht]
\centering
\begin{widetabular}{\columnwidth}{lcccccc}\toprule 
         & \multicolumn{6}{c}{$\mathrm{Error}_{\mathrm{IA}}$} \tabularnewline \cmidrule(lr){2-7} 
        H &    $w$=20  &  50  &  150 &  200 &  300 &  499 \tabularnewline
        \midrule  \relax 
        1024, 512      & 0.33 &0.08 &0.09 &0.10 &0.11 &2.53 \tabularnewline \relax  & (0.14) & (0.00) & (0.00) & (0.00) & (0.00) & (0.26) 
        \tabularnewline \relax 
        128, 128, 128   &   0.64 &0.08 &0.08 &0.07 & \textbf{0.06} &2.25 \tabularnewline \relax  & (0.11) & (0.01) & (0.00) & (0.00) & (0.00) & (0.18)
        \tabularnewline \bottomrule
\end{widetabular}
        \caption{Effect on $\mathrm{Error}_{\mathrm{IA}}$ of window $w$ and layer sizes $H$. }
\label{tab:w}
\end{table}

\paragraph{Relative error outperforms RMSE as a LIN loss function, longer sequence lengths can be helpful}
Optimizing the relative error loss ($L_{\mathrm{RE}}$) of $\widehat{l}$ rather than the RMSE significantly improved $\mathrm{Error}_{\mathrm{VF}}$ and led to the best  $\mathrm{Error}_{\mathrm{IA}}$ (Table \ref{tab:loss}). Use of $L_{\mathrm{RE}}$ as an optimization target may lead to better $\mathrm{Error}_{\mathrm{VF}}$ because each of these functions computes a relative error; i.e., $L_{\mathrm{RE}}$ has an advantage over RMSE when high relative error in latent space translates to high relative error in frame space. The fact that $L_{\mathrm{RE}}$ also led to the best $\mathrm{Error}_{\mathrm{IA}}$ may be attributable to the more accurate frame reconstructions obtained when using this loss.

Increasing the sequence length $s$ tended to improve $\mathrm{Error}_{\mathrm{VF}}$ (Table \ref{tab:loss}), regardless of optimization target (except with the transformer). Thus, for most LINs, to minimize discrepancies between CFD and AI surrogate simulations ($\mathrm{Error}_{\mathrm{VF}}$), our results support combining the relative error loss $L_{\mathrm{RE}}$ with $s>1$. Notably, \citet{kim2019deep} use RMSE and $s=1$ (with the MLP LIN).


\begin{table}[ht]
\centering
\begin{widetabular}{\columnwidth}{llcccc} 
\toprule             & {} & \multicolumn{2}{c}{$\mathrm{Error}_{\mathrm{IA}}$} & \multicolumn{2}{c}{$\mathrm{Error}_{\mathrm{VF}}$} \tabularnewline \cmidrule(lr){3-4} \cmidrule(lr){5-6}    
\relax
LIN & $s$ & $L_{\mathrm{RE}}$ & RMSE & $L_{\mathrm{RE}}$&  RMSE   \tabularnewline \midrule  \relax 
ARC&1 &0.10 (0.02) &0.09 (0.00) &0.52 (0.01) &0.55 (0.00) \tabularnewline \relax &6 &0.12 (0.03) &0.12 (0.02) &0.52 (0.01) &0.53 (0.00) \tabularnewline \relax LSTM&1 &0.11 (0.01) &0.09 (0.02) &0.53 (0.03) &0.54 (0.00) \tabularnewline \relax &6 &0.11 (0.01) &0.10 (0.01) &0.52 (0.00) &0.54 (0.00) \tabularnewline \relax MLP&1 & \textbf{0.06} (0.00) &0.08 (0.00) &0.53 (0.00) &0.55 (0.00) \tabularnewline \relax &6 &0.09 (0.00) &0.10 (0.03) & \textbf{0.49} (0.00) &0.52 (0.00) \tabularnewline \relax Transformer&1 &0.09 (0.01) &0.08 (0.04) &0.52 (0.00) &0.55 (0.00) \tabularnewline \relax &6 &0.09 (0.03) &0.14 (0.08) &0.53 (0.00) &0.68 (0.11) \tabularnewline
\bottomrule
\end{widetabular}
\caption{The effect on $\mathrm{Error}_{\mathrm{IA}}$ and $\mathrm{Error}_{\mathrm{VF}}$ of LIN architecture, input sequence length $s$, and training loss.}
\label{tab:loss}
\end{table}

\section{Path to Deployment}
\label{sec:deploy}
We now discuss the potential of our NN surrogate in the design of an industrial scale coal-fired power plant CCS and the necessary steps to achieve this deployment. At industrial scale, the 3D packed column for carbon capture is meters long, much larger than the 3D RCM shown in Figure \ref{fig:RCM}. To scale our approach to this case, two critical steps are required.  First, the surrogate will be modified to predict the data generated from the 3D RCM CFD simulations.  At this 3D RCM scale, the CFD simulation cost is manageable, so we can produce data as needed for training the NN surrogate.  Second, after verifying the accuracy of the 3D surrogates, they will be further scaled to predict industrial scale column data. Notably, the NN is well suited for this upscaling due to its remarkably cost-effective performance compared to the physics-based CFD simulations. Further, modifying our LVMs to accommodate 3D data is straightforward, and the knowledge and methodologies relevant to the 2D models can be transferred to the 3D case to avoid costly methodology optimization at the 3D scale.  However, we expect that the larger 3D domains will require more computing resources and data to reach acceptable performance levels.

Once trained, the 3D NN surrogates can provide data for various industry-scale CCS configurations via  high throughput IA prediction.  We can leverage this to perform IA-based CCS optimization by coupling the NNs with FOQUS \cite{eslick2014framework}, an advanced optimization and uncertainty quantification tool developed for carbon capture simulations.  Additionally, the NN surrogate-based IA prediction can be integrated into ASPEN \cite{van2020uncertainty}, which accounts for multiple systems in a coal-fired power plant to produce carbon capture techno-economic analysis.

\section{Discussion and Limitations}
IA-based design optimization of CCSs is prohibitively costly when relying on CFD, motivating our study of the DF surrogate simulation approach and  variants we introduced here. We found these AI surrogates could quickly simulate volume fraction field dynamics accurately enough to make low-error IA predictions for new CCS configurations. Indeed, DF appears to be promising for our application, as it is 4000x as fast as our CFD model with just 4\% IA error. 

While several of the DF surrogate variants we explored neared this performance level, our results on 2D data suggest that the most promising surrogate employs a CNN LVM, a transformer LIN, and E2E training. Thus, this surrogate will be our focus as we begin scaling to and evaluating on 3D CCS CFD data, a crucial next step on our path to  industrial-scale deployment of our DF-style surrogate models.

Importantly, though, DF has limitations for our application. The CNN LVM likely needs to be retrained to adapt to CCSs with different packings, as performance of CNNs on data distributions different from the train distribution significantly degrades \citep{hendrycks2018benchmarking,beery2018recognition}. 
For example, an LVM could minimize velocity field error by learning to always output 0 at packing locations, but doing so would cause high error on new data with different packing locations. To guard against this, an interesting future direction is to use graph neural networks (GNNs) to learn dynamics from nodes on graphs that represent mesh-based
simulation data. The GNN's learned internode dynamics have been shown to generalize well to new domains \citep{pfaff2020learning}, and such generalization could allow us to change CCS packings without retraining. 

Relatedly, our analysis left unclear whether performances observed on the data we studied would extend to models trained on new packing configurations or models tested on levels of $v$ beyond those seen during training. Notably, our analysis covered most practical operating values of $v$. Further, a preliminary study found that models that perform well on the packing configuration used here can be trained on a new packing configuration and obtain similar performance. Thus, the DF approach could potentially be used to train a model for each packing configuration, allowing the design-optimization procedure to quickly obtain IA values at any relevant $v$ via simulation with the model trained on the packing being considered (e.g., rather than using a single GNN model that works well for all packing configurations).

Another limitation of the DF approach is high field relative error. While errors of surrogate simulated fields were low enough to facilitate accurate IA estimates, similar applications may require more faithfulness to the CFD fields. Future work, akin to the end-to-end training we studied, could further improve relative errors of DF-predicted fields.


\section*{Acknowledgments}
This work was performed under the auspices of the U.S. Department of Energy by the Lawrence Livermore National Laboratory under Contract No. DE-AC52-07NA27344. The authors acknowledge financial support through the Point Source Carbon Capture Program at the U.S. Department of Energy, Office of Fossil Energy and Carbon Management. This work was conducted as part of the Carbon Capture Simulation for Industry Impact (CCSI2) project. We thank Phan Nguyen for helpful comments.

\bibliography{aaai22}
\bibliographystyle{plainnat}

\newpage
\appendix
\section{Appendix}
\label{sec:appendix}
\subsection{Additional model and training details}

We now describe additional hyperparameters related to the training and creation of models used in our experiments:

\begin{itemize}
    \item The ViT LVM and its linear LIN trained with batch size $b=16$. The CNN LVM trained with $b=128$.

    \item The ARC, LSTM, MLP, and transformer LINs trained with $b=32$. We also tried training LINs with batch sizes of 8, 64, and 128 but did not observe superior results. Notably, all LINs except for the transformer had consistent performance at different batch and training window sizes; for the transformer, $b=32$ and $w=150$ led to significantly better (and less variable) performance than other settings for these variables, which suggests that the training of the transformer may need to be further tuned. 

    \item When comparing the effect of different LIN loss functions and sequence lengths $s$ (see Table \ref{tab:loss}), we used $w=150$. 
    
    \item While larger values for $c$ led to better LVM results (Figure \ref{fig:LVM}), use of larger latent vectors makes the task of training the LIN more difficult \citep{kim2019deep}. Thus, when training the ARC, LSTM, MLP, and transformer LINs, we use $c=64$, as our preliminary experiments found good results with this setting.
    
    \item When performing end-to-end fine-tuning of the CNN LVM, ARC LIN surrogate  (see Table \ref{tab:e2e}), we used $b=32$, $w=499$, and initial learning rate $0.00001$. 
    
    \item The ViT LVM (see Table \ref{tab:e2e} and Figure \ref{fig:LVM}) was built with latent dimension $c=512$ and 3 encoder/decoder layers with 8 heads.
    
\end{itemize}   

All models (LVMs and LINs) were trained with the Adam optimizer \citep{kingma2014adam} and initial learning rate 0.001. For the CNN LVM and the ARC, LSTM, MLP, and transformer LINs, the learning rate was reduced to one-tenth its prior size upon plateau of the test loss (note: test loss was computed with the training window, rather than the window corresponding to the full rollout, which was used for comparison of model performance regardless of training window size). LVM learning rates were reduced after no improvement for six epochs, while LIN learning rates were reduced after no improvement for fifteen epochs (except when running experiments  to determine a good set of layer sizes for our LINs, see Table \ref{tab:w}, which reduced the learning rate after eight epochs of no improvement). The ViT LVM's learning rate and that of the corresponding linear LIN were reduced each epoch by five percent. 

Finally, note that our open-source code (\url{https://github.com/CCSI-Toolset/DeeperFluids}) is being released with a trained LIN and a trained LVM that can be used to run surrogate simulations. Relatedly, while the dataset of CFD simulation results we used to train our models is large (182GB when compressed), it can be made available upon request. 

\end{document}